\documentclass[10pt,twocolumn,letterpaper]{article}

\usepackage[pagenumbers]{cvpr} 

\usepackage{graphicx}
\usepackage{amsmath}
\usepackage{amssymb} 
\usepackage{booktabs}

\usepackage{flushend}

\usepackage[pagebackref,breaklinks,colorlinks]{hyperref}

\usepackage{float}
\usepackage{xcolor}
\usepackage{multicol}

\usepackage[capitalize]{cleveref}
\crefname{section}{Sec.}{Secs.}
\Crefname{section}{Section}{Sections}
\Crefname{table}{Table}{Tables}
\crefname{table}{Tab.}{Tabs.}

\def\cvprPaperID{16} 
\def\confName{CV4Animals}
\def\confYear{2023}

\begin{document}

\title{Predictive Modeling of Equine Activity Budgets Using a 3D Skeleton Reconstructed from Surveillance Recordings}







\author{Ernest Pokropek$^1$ ~~~~ 
Sofia Broom{\'e}$^{1,2}$ ~~~~ Pia Haubro Andersen$^3$ ~~~~ Hedvig Kjellstr{\"o}m$^{1,3}$\\
$^1$ KTH, Sweden {\tt pokropek,sbroome,hedvig@kth.se} ~~~~
$^2$ Therapanacea, France \\
$^3$ SLU, Sweden {\tt pia.haubro.andersen@slu.se}}


\maketitle

\begin{abstract}
In this work, we present a pipeline to reconstruct the 3D pose of a horse from 4 simultaneous surveillance camera recordings. Our environment poses interesting challenges to tackle, such as limited field view of the cameras and a relatively closed and small environment. The pipeline consists of training a 2D markerless pose estimation model to work on every viewpoint, then applying it to the videos and performing triangulation. We present numerical evaluation of the results (error analysis), as well as show the utility of the achieved poses in downstream tasks of selected behavioral predictions. Our analysis of the predictive model for equine behavior showed a bias towards pain-induced horses, which aligns with our understanding of how behavior varies across painful and healthy subjects.
\end{abstract}


%

\section{Introduction}
\label{sec:intro}

Animal welfare science is increasingly focusing on not only assessing the quality of the environment in which animals are kept but also evaluating their physical and psychological well-being. In this context, behavior is a valuable indicator of welfare and disease \cite{price_preliminary_2003, ashley_behavioural_2005, hausberger_detecting_2016}, providing insight into the animal's subjective state. One way to assess behavior is through the evaluation of the amount of time spent engaging in different activities, also known as a time budget. Changes in a time budget can help identify painful conditions and evaluate the effectiveness of management interventions to improve equine welfare. However, measuring time budgets requires detailed and lengthy surveillance, which is usually done through direct observation by human observers. This method is prone to bias, fatigue, and limited accuracy and temporal resolution. Additionally, direct observation and video analysis are resource-intensive, which restricts the observation periods and the number of individuals studied. Thus, for time budget analysis to become a reliable and useful tool for welfare assessment, automated recognition of behavior is necessary \cite{auer_activity_2021}.

Pose estimation employs machine learning models to estimate the positions of points/areas of interest of a given agent (or multiple agents) in video data, for them to be later connected creating a pose. This can be performed to produce an either 2-dimensional 2D or 3-dimensional 3D result, most commonly used with human agents \cite{Zheng2019DeepLH}. When it comes to the latter, the methodology usually consists of first performing the 2D pose estimation from various angles and then combining the results in the process called \textit{lifting}, which commonly involves various triangulation techniques. Pose reconstruction is especially useful in areas of augmented reality \cite{7368948}, healthcare \cite{s21217315}, motion analysis \cite{DESMARAIS2021103275} and many more, vastly reducing the necessity of computational and financial resources and outperforming previous detection methods \cite{Zheng2019DeepLH}. Furthermore, it showed a distinct improvement over commonly used methodologies for tasks such as object detection \cite{NIPS2015_14bfa6bb} or image classification \cite{NIPS2012_c399862d}.

 \begin{figure*}[h]
  \centerline{\includegraphics[width=\textwidth]{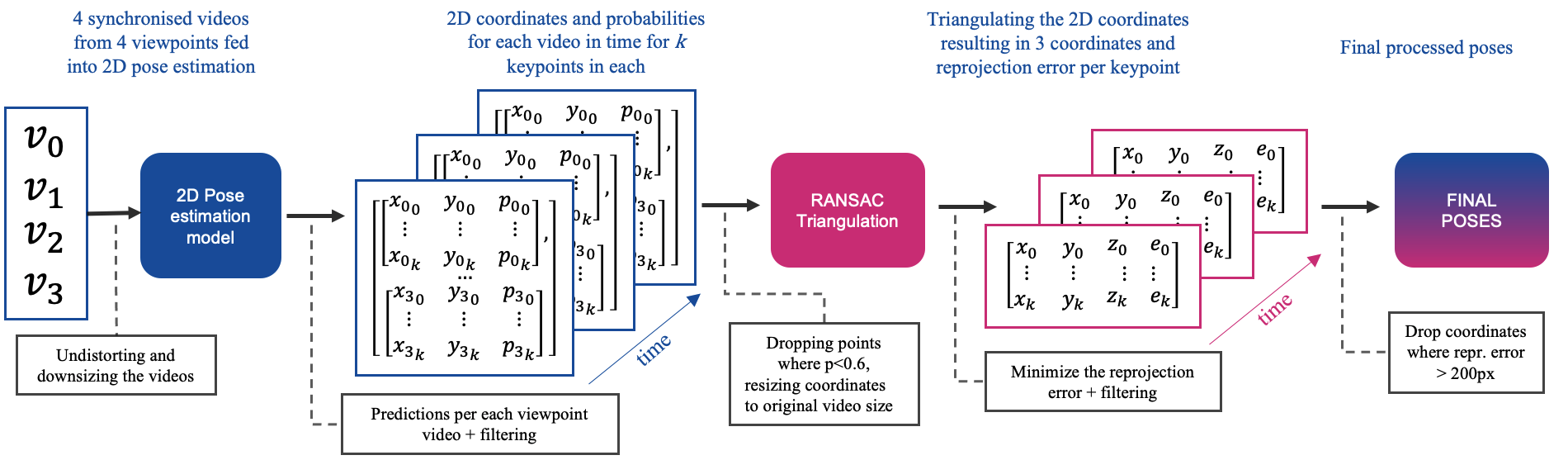}}
    \vspace{-1mm}

  \caption{Visualized proposed pipeline for 3D pose estimation of horses upon receiving new data.}
  \label{fig:pipeline}
\end{figure*}

In this work, we present a pipeline for reconstructing the 3D pose of a whole horse constructed using 28 keypoints in a challenging environment. We show the utilization of the obtained skeleton in a downstream task of behavior/activity prediction. Our pipeline addresses the 3D pose estimation task by solving two major challenges: (1) 2D keypoint estimation and (2) lifting the 2D keypoints into 3 dimensions. The first part was implemented using the DeepLabCut \cite{Mathisetal2018} interface, where we used pretrained ResNet50 weights and hand labelled data to fit a single keypoint estimation model for all viewpoints. For the latter, we used functions provided from the \texttt{aniposelib} library \cite{karashchuk_anipose_2020} and performed a RANSAC triangulation on keypoints estimated on videos from 4 viewpoints.




\section{Related work}
\label{sec:relwork}

3D pose reconstruction has been widely used for humans and it also gained popularity with animal data. Nath et al. provide an exhaustive description on marker-less pose estimation for various animals and they present an example of 3D reconstruction for cheetahs in the wild from 6 camera recordings using the DeepLabCut interface \cite{NathMathis2018}.
One of the potential use cases of such reconstruction is a pipeline involving object detection, 2D pose estimation as well as 3D reconstruction generates pose of the animal to be later used for wildlife animation\cite{Fangbemi2020ZooBuilder2A}. 
Without a surprise, the scientific community noticed many problems regarding the unpredictability of animal data and resources needed for triangulation.
With concerns about the unpredictability of animal data and resources needed for triangulation, 
a framework called LiftPose3D is introduced, which via deep neural networks estimates the 3D pose of an animal based on a single camera view \cite{gosztolai_liftpose3d_2021}. This work is especially important given that sometimes one may not have access to many views of the animal, as well as to the camera calibration parameters, making the standard triangulation unfeasible. Furthermore, datasets dedicated to
animal pose estimation have been introduced, where a domain adaptation approach is used to reconstruct the skeleton, achieving results as good as for the more explored
human pose estimation \cite{9009505}.

DeepFly3D \cite{Gunel19DeepFly3D} uses deep neural networks to estimate the 3D pose of tethered Drosophila melanogaster. It eliminates the need for manual calibration, incorporates error detection and correction, and improves performance through active learning. DeepFly3D enables detailed automated behavioral measurements for various biological applications.

Action segmentation has also been employed in animal behavior analysis. SIPEC \cite{Marks2020DeeplearningBI}, a novel deep learning architecture, enables the classification of individual and social animal behavior in complex environments directly from raw video frames. It outperforms existing methods and successfully recognizes behaviors of freely moving mice and socially interacting non-human primates using simple mono-vision cameras in home-cage setups.

\section{Data and methods}
\label{sec:data_methods}
\begin{figure}[h]
  \includegraphics[width=0.473\textwidth]{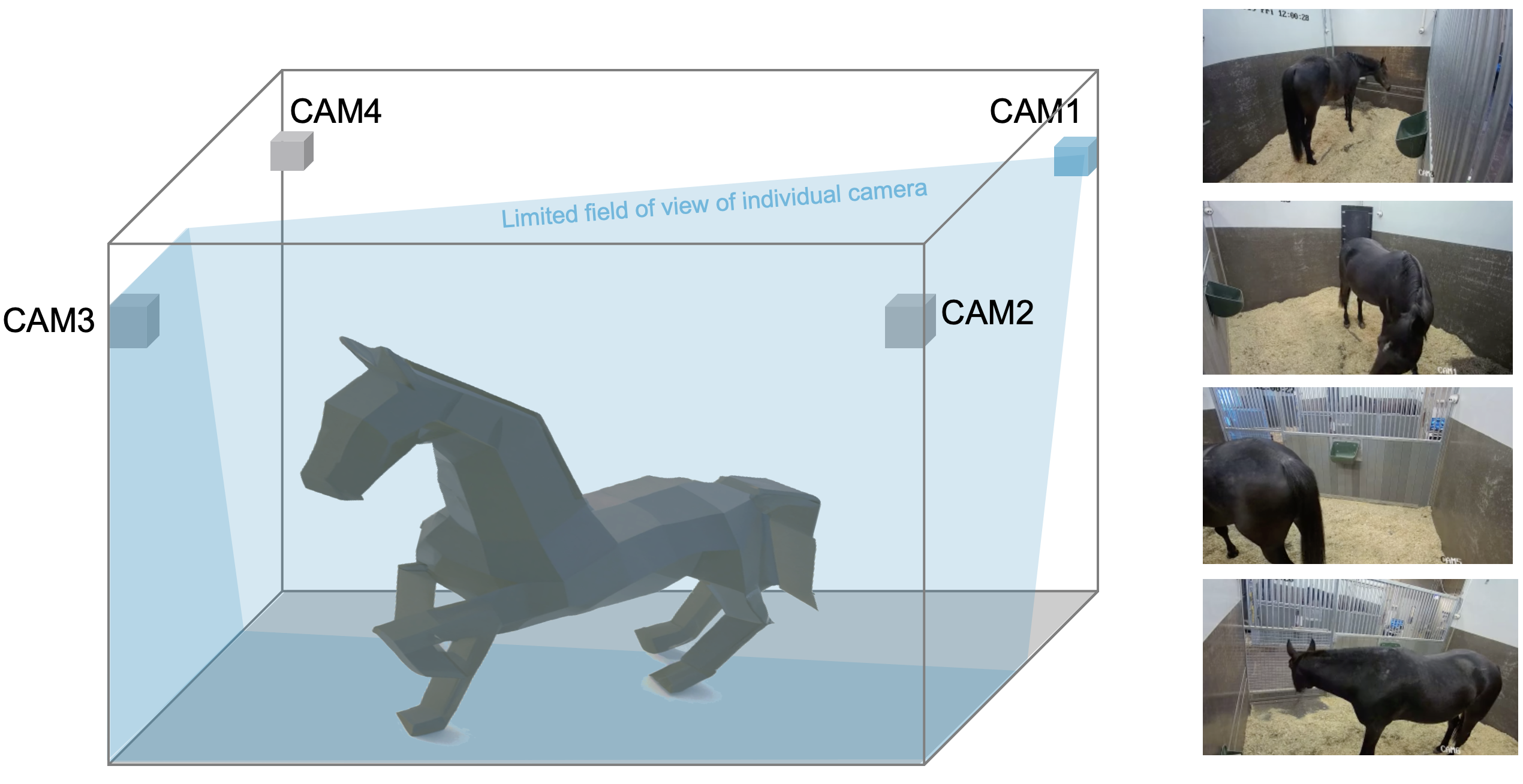}
  \vspace{-1mm}
  \caption{Environment in the data for this study. Please note that each individual camera has a limited field of view and often does not capture the whole animal.}
  \label{fig:setup}
\end{figure}

The workflow of this study is illustrated in Figure \ref{fig:pipeline}. Data for veterinary research has been used; it consists of approximately 45 minute recordings (approximately 20 frames per second each) of horses in rectangular enclosures, filmed simultaneously from 4 viewpoints placed in the corners, above the horse \cite{Ask2020}. Due to noise in the data set, we used 13 subjects with reliable videos, resulting in 52 videos being analyzed in this study. Some of the subjects (9) were exposed to pain, which affected their behavior. Visualization, together with examples from the real dataset, can be seen in Figure \ref{fig:setup}. This data set is quite challenging - each camera has a limited field of view, with parts of the animal not being visible from certain angles.

Videos have been undistorted (camera distortions have been removed) as well as downsized to resolution of 336x190. The horses participated in an orthopedic pain study and videos were recorded before the horses were in pain and during pain. All videos were manually annotated for all observable state and point events, according to an ethogram \cite{noauthor_aktivitetsbudget_nodate}.
Activity budgets were available for basic behaviors for horses in no-pain and pain states.

\begin{figure*}[t]
 \vspace{-5mm} \centerline{\includegraphics[width=\textwidth]{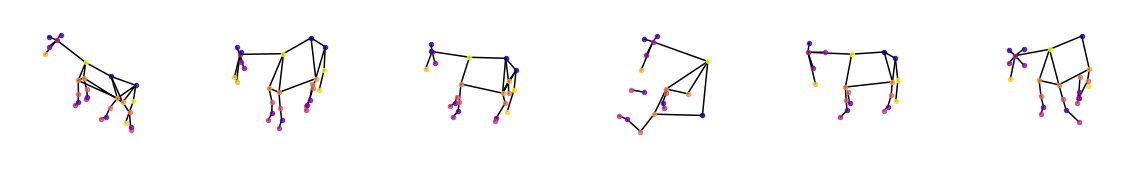}}
  \vspace{-6mm} 
  \caption{Example 3D poses of the horse. 
 Note that our pipeline was 
 able to also depict more unconventional poses such as the fourth one 
  (the horse was sitting).
  The disconnected joints are the result of missing keypoints, dropped either due to low probability (during 2D estimation) or high reprojection error (during 3D lifting).
  Different angles correspond to different orientations of the animal in the box.}
  \label{fig:poses}
\end{figure*}

\subsection{Automated keypoint estimation}
In this study, 28 keypoints were selected to represent the pose of the horse in a simplistic but informative manner. First, one has to manually annotate a portion of the frames; in case of this study, it has been performed using the open-source marker-less pose estimation package DeepLabCut designed for animal data \cite{Mathisetal2018} in its Python implementation \cite{NathMathisetal2019}. In total, we extracted 1550 frames from the videos: some sampled randomly, some sampled via k-means that maximize the generalization capabilities of the frames, and lastly via sampling 'outlier' frames. When the
initial model was trained, we selected the frames that the model was performing poorly on and labelled them to later merge it into the training set, as proposed by the DeepLabCut method. The model with ResNet50 weights has been trained for 500,000 iterations, until convergence.
It was then used to predict the keypoints for whole length of videos, and the predictions were passed through an arima filter \cite{arima}. Finally, we dropped the points with probability smaller than 0.6, in order to remove unsure predictions.

\subsection{3D pose reconstruction}
Having access to many cameras placed around the agent with knowledge of their calibration parameters, it is possible to lift 
the set of 2D coordinates into 3D. 
In this study, it has been done via the RANSAC triangulation algorithm using the Anipose package for Python \cite{karashchuk_anipose_2020} which provides a triangulation interface for many viewpoints and is directly compatible with DeepLabCut. After receiving the 2D keypoint estimation for all videos from the previous step, we resize these coordinates to correspond to the original video size (due to having the calibration parameters corresponding to that). Then, we perform the triangulation while minimizing the reprojection error (we project the 3D coordinates to 2D, calculate the error of projection, and repeat that for all viewpoints). Additional filtering (median filter) of the final pose is added to smooth out the noise, and for the final poses we drop the individual keypoints for which the 3D reprojection error is bigger than 200 pixels, which we found to be a reasonable threshold four our data that does allow for some level of uncertainty. Not removing these keypoints would in best case scenario introduce unnecessary noise for downstream task of behavior prediction, and in worst, potentially bias the downstream prediction model to identify subjects based on this noise. Usually, high reprojection error was caused by the keypoint not being properly visible from some of the viewpoints, which could indicate the animal's placement within the environment, which varied across the subjects.

\subsection{Downstream prediction tasks}

To further evaluate the quality of our 3D pose estimation, we have used the available annotations of horses' activity budgets and performed a binary classification for them, using only skeleton data. Because movement, eating, and standing are essential behaviors in an activity time budget and have been found to exhibit varying durations in horses experiencing pain versus those that are not, these were chosen for analysis based on an ethogram \cite{noauthor_aktivitetsbudget_nodate}. Data from 13 horses has been used, with videos of 
$\sim$ 45 minutes each. Although given the dense nature of the behavior annotations, action segmentation could have been used, we opted for action recognition frameworks given their popularity and relatively lower computational cost, as the presented pipeline should be able to operate in real time.

\vspace{-2mm}
\subsubsection{Data preparation}
\vspace{-1mm}
For each annotated interval of the behavior happening, we randomly selected a 200 frame segment of it. Since we perform this experiment in a binary fashion, for each such segment, we also picked another one (of same length, from same subject), where the behavior was not occurring. For each horse, if there was more than 5 such pairs, we randomly selected 1 segment pair for the validation set and 3 pairs for the test set (without replacement). The remaining segments were used for training. This resulted in 986 annotated segments (200 frames each) used for training, 58 for validation, and 174 for test. Each segment corresponds to one label to predict.

\vspace{-2mm}
\subsubsection{Training and evaluation}
\vspace{-1mm}
We performed two main experiments: (1) training a Random Forest Classifier (RFC) with plain 3D coordinates (using 100 tree estimators), and (2) training a Spatial Temporal Graph Convolutional Network (ST-GCNN), which leverages temporal information and has been successfully applied to human skeleton data for activity recognition \cite{Yan2018SpatialTG}. In order to produce one prediction per each 200 frame segment, validation for the RFC was conducted in a voting manner. Prediction for each frame was produced, and then they were averaged across all frames. The resulting value was then rounded.

Each ST-GCNN was trained for 50 epochs, using the parameters from its original implementation. Both models were trained to perform binary classification for the selected 3 activities (one model each per behavior). For both experiments, the missing values (due to high reprojection error) were replaced with zeros.


\section{Results}
\label{sec:results}

\begin{table}[]

\begin{tabular}{rrrr}
\textbf{}                                     & \textbf{Repr. Err. {[}px{]}} & \textbf{KPP}  & \textbf{KP\textsubscript{no}}    \\ \cline{2-4} 

\multicolumn{1}{r|}{\textbf{Nostrils}}          & \multicolumn{1}{r|}{49.41±50.43}     & \multicolumn{1}{r|}{47.55\%}  & \multicolumn{1}{r|}{2} \\ \cline{2-4} 
\multicolumn{1}{r|}{\textbf{Ears}}              & \multicolumn{1}{r|}{108.45±44.11}    & \multicolumn{1}{r|}{66.71\%}  & \multicolumn{1}{r|}{2}\\ \cline{2-4} 
\multicolumn{1}{r|}{\textbf{Eyes}}              & \multicolumn{1}{r|}{88.03±46.35}     & \multicolumn{1}{r|}{56.275\%} & \multicolumn{1}{r|}{2} \\ \cline{2-4} 
\multicolumn{1}{r|}{\textbf{Head Top}}              & \multicolumn{1}{r|}{112.59±44.85}    & \multicolumn{1}{r|}{63.41\%}  & \multicolumn{1}{r|}{1}\\ \cline{2-4} 
\multicolumn{1}{r|}{\textbf{Withers}}               & \multicolumn{1}{r|}{110.31±42.00}    & \multicolumn{1}{r|}{76.78\%}  & \multicolumn{1}{r|}{1}\\ \cline{2-4} 
\multicolumn{1}{r|}{\textbf{Croup}}                 & \multicolumn{1}{r|}{113.66±40.05}    & \multicolumn{1}{r|}{56.39\%}  & \multicolumn{1}{r|}{1}\\ \cline{2-4} 
\multicolumn{1}{r|}{\textbf{Tail}}              & \multicolumn{1}{r|}{80.87±45.45}     & \multicolumn{1}{r|}{52.87\%}  & \multicolumn{1}{r|}{3} \\ \cline{2-4} 
\multicolumn{1}{r|}{\textbf{Legs}}  & \multicolumn{1}{r|}{83.78±49.75}     & \multicolumn{1}{r|}{70.17\%}  & \multicolumn{1}{r|}{16} \\ \cline{2-4} 
\textbf{}    
\end{tabular}
\vspace{-5mm}
\caption{Reprojection errors 
for groups of keypoints in 3D for all 
poses. KPP stands for Key Point Presence, that is how often these keypoints were on average present (not dropped because of high reprojection errors). KP\textsubscript{no} is the number of keypoints in the group (for example, 2 keypoints for ears since there is one per 
side).}
\label{table:keypoints}
\end{table}

In total, we estimated over 700,000 poses with mean reprojection error of 89.90$\pm$45.98px (on an original video size of 2688x1520), with median reprojection error being slightly lower (84.60px). On average, each pose consists of 18$\pm$3 3D keypoints (out of total 28). Example poses are shown in Figure \ref{fig:poses} and more detailed results (per 
keypoint group) are presented in Table \ref{table:keypoints}. The errors per 
keypoint and the missing rate per keypoint tend to fluctuate around similar magnitudes. 


\begin{figure}[h]
    \centering
    \includegraphics[width=0.5\textwidth]{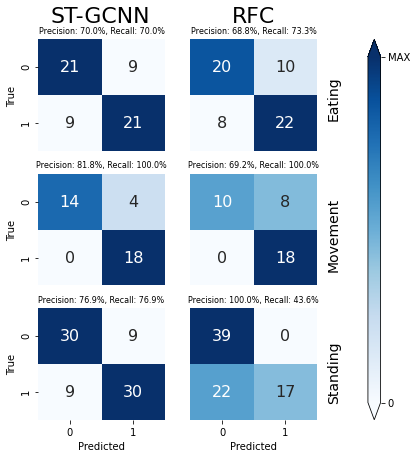}
 \vspace{-6mm}
    \caption{Confusion matrices for each behavior rows 
    and model columns 
    evaluated on the test set (each prediction corresponds to a 200 frame-long segment). 1 corresponds to the activity happening, whereas 0 to not happening. Precision and recall computed for each confusion matrix annotated on top of each matrix. 
    }
    \label{fig:behavior_cms}
\end{figure}

Basing only on the 3D coordinates, RFC was able to correctly classify eating behavior, while struggling with movement and standing, as there was no information about the dynamics of the skeleton.
When utilizing the temporal information, the predictive capabilities of the model using the animal's pose are especially well-performing. 
 This can be seen 
in Figure \ref{fig:behavior_cms} with both precision and recall oscillating above 70\% for ST-GCNN for all behaviors. As expected, eating is the easiest to predict on static data, as it is usually represented by the horse lowering its head.

\begin{table}[h]
\begin{tabular}{rcc||cc}
\multicolumn{1}{l}{}                    & \multicolumn{2}{c}{\begin{tabular}[c]{@{}c@{}}\textbf{Prediction}\\\textbf{distribution}\end{tabular}} & \multicolumn{2}{c}{\textbf{Performance}}                     \\ \cline{2-5} 
\multicolumn{1}{l|}{\textbf{}}          & \multicolumn{1}{c|}{Healthy}                & \multicolumn{1}{c||}{Painful}               & \multicolumn{1}{c|}{Precision} & \multicolumn{1}{c|}{Recall} \\ \hline
\multicolumn{1}{|r|}{\textbf{Eating}}   & \multicolumn{1}{c|}{0.56}                   & \multicolumn{1}{c||}{0.61}               & \multicolumn{1}{c|}{52.4\%}    & \multicolumn{1}{c|}{61.1\%} \\ \hline
\multicolumn{1}{|r|}{\textbf{Movement}} & \multicolumn{1}{c|}{0.42}                   & \multicolumn{1}{c||}{0.67}               & \multicolumn{1}{c|}{69.2\%}    & \multicolumn{1}{c|}{75.0\%} \\ \hline
\multicolumn{1}{|r|}{\textbf{Standing}} & \multicolumn{1}{c|}{0.43}                   & \multicolumn{1}{c||}{0.73}               & \multicolumn{1}{c|}{71.4\%}    & \multicolumn{1}{c|}{83.3\%} \\ \hline
\end{tabular}
\vspace{-1mm}
\caption{Prediction distribution of behaviors for healthy and painful subjects in the test set (trained using ST-GCNN on pain-balanced data), normalized by the total number of predictions per each behavior for healthy/painful group. 0.56 for 'Eating' behavior for healthy subjects means that the model predicted eating for 56\% of the segments involving healthy horses.}
\label{tab:prediction_distribution}
\end{table}


We conducted an experiment to test for predictive bias towards pain behavior in horses. The original training data had imbalances in terms of pain subjects, with some segments having nearly triple the instances, so we balanced it by downsampling the majority class. We trained ST-GCNN on this data and tested against the unchanged test set. Achieving good metrics was difficult due to the significant reduction in the training dataset size.

Table \ref{tab:prediction_distribution} presents the results of this experiment. Effectively, we have investigated the prediction distribution of the models that were trained on 
an equal amount of segments with healthy and painful subjects. 
The smallest difference in those distributions is attributed to eating, however, given the small amount of data, we were not able to train this classifier to achieve satisfactory performance. Nevertheless, when performing better than at random, it exhibits the behaviors expected for these subjects. In total, there is a bias for false positives to be predicted for the painful group. For more details about how this
compares to the distribution of actual behaviors, see Section \ref{sec:conclusion}.

\section{Discussion}
\label{sec:conclusion}
In this work, we presented our pipeline for reconstructing the 3D pose of a horse in challenging environments. Apart from the investigation of the errors, we presented a downstream task as an example of how these coordinates may become useful for predictive challenges. However, there are some improvements that could have been made to this work: in order for the triangulation to work as intended, the videos from different viewpoints need to be well synchronized. Unfortunately, in case of our data it was not always the case, and it occurred in a non-systematic manner, hence fixing it became an endeavour on its own. Also, having a lot of missing coordinates in 3D (due to various reasons), it may make sense to interpolate them based on previous and proceeding values for a more continuous and stable pose reconstruction. 

The behavior prediction could also be improved. The data amount used is small ompared to human pose action recognition datasets; when utilizing more data, the performance metric would probably improve significantly, given the good performance on the current dataset.

It should further be noted that this environment is incredibly challenging -- most of the time, some 
keypoints on the horse are occluded. Usually, the best case scenario is having a point visible from 3 cameras, and even that rarely happens -- the horse would need to place itself in the middle of the box, ideally with the head up.

The proposed method for automated determination of an activity budget has demonstrated promising potential, as evidenced by the results obtained shown in Figure \ref{fig:behavior_cms} and Table \ref{tab:prediction_distribution}. All selected behaviors exhibited the same variation as earlier detected by manual observation and labeling, as reported by Pålsson. Interestingly, the behavior "eating" revealed that the painful horses had more eating time than the sound horses, which is controversial, since painful horses are expected not to eat \cite{hausberger_detecting_2016}. However, the Pålsson study also indicated that painful horses had more eating behavior, which was interpreted as a comforting behavior. Further research is needed to differentiate between "eating" and the head position "down". The changes in "movement" indicate that painful horses tend to be more restless, which aligns with previous findings by Ashley and Hausberger. Moreover, the study revealed that painful horses are less inclined to lie down, as shown in Table \ref{tab:prediction_distribution}. 

It has to be noted, that the prediction of behavior is not perfect and shows notable errors, which could undermine the statistical significance of results presented in Table \ref{tab:prediction_distribution}. The direction for future studies would be to analyse the bias of the predictive model towards pain assessment when training data is balanced and yields good validation results. 

Despite the study's limitations, our method shows great potential for the automated recognition of 
horse behaviors and their time budgets.

{\small
\bibliographystyle{ieee_fullname}
\bibliography{bibliography}

\begin{thebibliography}{10}\itemsep=-1pt

\bibitem{arima}
Nari Arunraj, Diane Ahrens, and Michael Fernandes.
\newblock Application of sarimax model to forecast daily sales in food retail
  industry.
\newblock {\em International Journal of Operations Research and Information
  Systems}, 7:1--21, 04 2016.

\bibitem{ashley_behavioural_2005}
F.~H. Ashley, A.~E. Waterman-Pearson, and H.~R. Whay.
\newblock Behavioural assessment of pain in horses and donkeys: application to
  clinical practice and future studies.
\newblock {\em Equine Veterinary Journal}, 37(6):565--575, Nov. 2005.

\bibitem{Ask2020}
Katrina Ask, Marie Rhodin, Lena-Mari Tamminen, Elin Hernlund, and Pia
  Haubro~Andersen.
\newblock Identification of body behaviors and facial expressions associated
  with induced orthopedic pain in four equine pain scales.
\newblock {\em Animals}, 10(11), 2020.

\bibitem{auer_activity_2021}
Ulrike Auer, Zsofia Kelemen, Veronika Engl, and Florien Jenner.
\newblock Activity {Time} {Budgets}—{A} {Potential} {Tool} to {Monitor}
  {Equine} {Welfare}?
\newblock {\em Animals}, 11(3):850, Mar. 2021.
\newblock Number: 3 Publisher: Multidisciplinary Digital Publishing Institute.

\bibitem{9009505}
Jinkun Cao, Hongyang Tang, Hao-Shu Fang, Xiaoyong Shen, Yu-Wing Tai, and Cewu
  Lu.
\newblock Cross-domain adaptation for animal pose estimation.
\newblock In {\em 2019 IEEE/CVF International Conference on Computer Vision
  (ICCV)}, pages 9497--9506, 2019.

\bibitem{DESMARAIS2021103275}
Yann Desmarais, Denis Mottet, Pierre Slangen, and Philippe Montesinos.
\newblock A review of 3d human pose estimation algorithms for markerless motion
  capture.
\newblock {\em Computer Vision and Image Understanding}, 212:103275, 2021.

\bibitem{Fangbemi2020ZooBuilder2A}
Abassin~Sourou Fangbemi, Yi~Fei Lu, Mao~Yuan Xu, Xiaoming Luo, Alexis Rolland,
  and Chedy Ra{\"i}ssi.
\newblock Zoobuilder: 2d and 3d pose estimation for quadrupeds using synthetic
  data.
\newblock {\em ArXiv}, abs/2009.05389, 2020.

\bibitem{gosztolai_liftpose3d_2021}
Adam Gosztolai, Semih Günel, Victor Lobato-Ríos, Marco Pietro~Abrate, Daniel
  Morales, Helge Rhodin, Pascal Fua, and Pavan Ramdya.
\newblock {LiftPose3D}, a deep learning-based approach for transforming
  two-dimensional to three-dimensional poses in laboratory animals.
\newblock {\em Nature Methods}, 18(8):975--981, Aug. 2021.
\newblock Number: 8 Publisher: Nature Publishing Group.

\bibitem{Gunel19DeepFly3D}
Semih G{\"u}nel, Helge Rhodin, Daniel Morales, João Compagnolo, Pavan Ramdya,
  and Pascal Fua.
\newblock Deepfly3d, a deep learning-based approach for 3d limb and appendage
  tracking in tethered, adult drosophila.
\newblock In {\em eLife}, 2019.

\bibitem{hausberger_detecting_2016}
Martine Hausberger, Carole Fureix, and Clémence Lesimple.
\newblock Detecting horses’ sickness: {In} search of visible signs.
\newblock {\em Applied Animal Behaviour Science}, 175:41--49, 2016.

\bibitem{karashchuk_anipose_2020}
Pierre Karashchuk, Katie~L. Rupp, Evyn~S. Dickinson, Elischa Sanders, Eiman
  Azim, Bingni~W. Brunton, and John~C. Tuthill.
\newblock Anipose: a toolkit for robust markerless {3D} pose estimation.
\newblock Technical report, bioRxiv, May 2020.
\newblock Section: New Results Type: article.

\bibitem{NIPS2012_c399862d}
Alex Krizhevsky, Ilya Sutskever, and Geoffrey~E Hinton.
\newblock Imagenet classification with deep convolutional neural networks.
\newblock In F. Pereira, C.~J.~C. Burges, L. Bottou, and K.~Q. Weinberger,
  editors, {\em Advances in Neural Information Processing Systems}, volume~25.
  Curran Associates, Inc., 2012.

\bibitem{7368948}
Eric Marchand, Hideaki Uchiyama, and Fabien Spindler.
\newblock Pose estimation for augmented reality: A hands-on survey.
\newblock {\em IEEE Transactions on Visualization and Computer Graphics},
  22(12):2633--2651, 2016.

\bibitem{Marks2020DeeplearningBI}
Markus Marks, Qiuhan Jin, Oliver Sturman, Lukas~M. von Ziegler, Sepp
  Kollmorgen, Wolfger von~der Behrens, Valerio Mante, Johannes Bohacek, and
  Mehmet~Fatih Yanik.
\newblock Deep-learning based identification, tracking, pose estimation, and
  behavior classification of interacting primates and mice in complex
  environments.
\newblock {\em Nature machine intelligence}, 4:331 -- 340, 2020.

\bibitem{Mathisetal2018}
Alexander Mathis, Pranav Mamidanna, Kevin~M. Cury, Taiga Abe, Venkatesh~N.
  Murthy, Mackenzie~W. Mathis, and Matthias Bethge.
\newblock Deeplabcut: markerless pose estimation of user-defined body parts
  with deep learning.
\newblock {\em Nature Neuroscience}, 2018.

\bibitem{NathMathis2018}
Tanmay Nath*, Alexander Mathis*, An~Chi Chen, Amir Patel, Matthias Bethge, and
  Mackenzie~W Mathis.
\newblock Using deeplabcut for 3d markerless pose estimation across species and
  behaviors.
\newblock {\em bioRxiv}, 2018.

\bibitem{NathMathisetal2019}
Tanmay Nath*, Alexander Mathis*, An~Chi Chen, Amir Patel, Matthias Bethge, and
  Mackenzie~W Mathis.
\newblock Using deeplabcut for 3d markerless pose estimation across species and
  behaviors.
\newblock {\em Nature Protocols}, 2019.

\bibitem{price_preliminary_2003}
Jill Price, Seago Catriona, Elizabeth~M. Welsh, and Natalie~K. Waran.
\newblock Preliminary evaluation of a behaviour-based system for assessment of
  post-operative pain in horses following arthroscopic surgery.
\newblock {\em Veterinary Anaesthesia and Analgesia}, 30(3):124--137, July
  2003.

\bibitem{noauthor_aktivitetsbudget_nodate}
Linnéa Pålsson.
\newblock Activity budgets and pain related behaviours in horses with induced
  orthopedic pain, 2020.

\bibitem{NIPS2015_14bfa6bb}
Shaoqing Ren, Kaiming He, Ross Girshick, and Jian Sun.
\newblock Faster r-cnn: Towards real-time object detection with region proposal
  networks.
\newblock In C. Cortes, N. Lawrence, D. Lee, M. Sugiyama, and R. Garnett,
  editors, {\em Advances in Neural Information Processing Systems}, volume~28.
  Curran Associates, Inc., 2015.

\bibitem{s21217315}
Jan Stenum, Kendra~M. Cherry-Allen, Connor~O. Pyles, Rachel~D. Reetzke,
  Michael~F. Vignos, and Ryan~T. Roemmich.
\newblock Applications of pose estimation in human health and performance
  across the lifespan.
\newblock {\em Sensors}, 21(21), 2021.

\bibitem{Yan2018SpatialTG}
Sijie Yan, Yuanjun Xiong, and Dahua Lin.
\newblock Spatial temporal graph convolutional networks for skeleton-based
  action recognition.
\newblock In {\em AAAI Conference on Artificial Intelligence}, 2018.

\bibitem{Zheng2019DeepLH}
Ce Zheng, Wenhan Wu, Taojiannan Yang, Sijie Zhu, Chen Chen, Ruixu Liu, Ju Shen,
  Nasser Kehtarnavaz, and Mubarak Shah.
\newblock Deep learning-based human pose estimation: A survey.
\newblock {\em ArXiv}, abs/2012.13392, 2019.

\end{thebibliography}
}

\end{document}